\documentclass[runningheads]{llncs}

\usepackage{eccv}

\usepackage{eccvabbrv}

\usepackage{graphicx}
\usepackage{booktabs}

\usepackage[accsupp]{axessibility}

\usepackage{hyperref}

\usepackage{eso-pic}

\usepackage{orcidlink}

\graphicspath{{figs/}}

\begin{document}

\AddToShipoutPictureFG*{\AtTextLowerLeft{\put(0,-15){\parbox[t]{\textwidth}{\scriptsize\raggedright
This version of the contribution has been accepted for publication, after peer review (when applicable) but is not the Version of Record and does not reflect post-acceptance improvements, or any corrections. The Version of Record is available online at: \url{http://dx.doi.org/10.1007/978-3-032-37356-4_34}. Use of this Accepted Version is subject to the publisher's Accepted Manuscript terms of use \url{https://www.springernature.com/gp/open-research/policies/accepted-manuscript-terms}.}}}}

\title{NanoVSR: Towards Real-Time Video Super-Resolution on Edge Devices} 

\titlerunning{NanoVSR: Towards Real-Time VSR on Edge Devices}

\author{Filip Pawlicki \and
Marcel Kańduła \and
Marcin Pucek \and
Kamil Dobies}
\authorrunning{F.~Pawlicki et al.}

\institute{Gdańsk University of Technology\\
\email{filip.pawlicki@nask.pl}}

\maketitle

\begin{abstract}
Recent Video Super-Resolution (VSR) methods rely heavily on transformers and explicit optical flow, creating computational overhead and custom operations that hinder deployment on hardware accelerators like TensorRT. To address this, we introduce NanoVSR, a scalable, fully convolutional architecture designed for resource-constrained edge devices. Using structural reparameterization, NanoVSR collapses into standard convolutions during inference, ensuring seamless hardware compatibility and negligible runtime overhead. Furthermore, despite lacking explicit motion compensation, it maintains competitive restoration quality by implicitly learning spatio-temporal alignments through progressive training. Evaluated on the REDS4 benchmark, NanoVSR demonstrates an exceptional balance between accuracy and computational efficiency, significantly improving the trade-off for compact architectures. Our NanoVSR-644k baseline yields 28.64 dB PSNR while delivering 27.2 FPS on the NVIDIA Jetson Orin NX 16GB (25W), offering massive speed gains over heavier models. The scaled NanoVSR-1.7M variant reaches 29.15~dB with a throughput of 19.58~FPS, providing superior, edge-optimized upscaling. Code is available at \url{https://github.com/filippawlicki/nanovsr}.
\end{abstract}

\section{Introduction}
\label{sec:intro}

Super-resolution (SR) is a fundamental challenge in computer vision aiming to reconstruct high-resolution (HR) images from low-resolution (LR) counterparts. While single image super-resolution focuses on recovering missing high-frequency details from a static image~\cite{dong2016imagesuperresolutionusingdeep, zhang2018imagesuperresolutionusingdeep}, it is inherently limited by the absence of external information. Video super-resolution (VSR) transcends this limitation by introducing the temporal dimension, leveraging the rich, redundant information scattered across consecutive frames to recover details that are physically absent in a single frame. The fundamental mechanism of VSR involves the effective aggregation of this temporal data; however, because objects and cameras move, simply stacking neighboring frames results in ghosting and blurring. Consequently, the core challenge of VSR lies in finding effective methods for motion compensation.

Recent studies~\cite{wang2019edvr, chan2021basicvsr, liang2022recurrent} have produced architectures capable of achieving impressive visual quality, successfully recovering details and maintaining temporal consistency across frames. To achieve this high standard of restoration, contemporary methods typically employ very deep neural networks that process video sequences using bidirectional propagation. Crucially, they rely on sophisticated alignment modules, utilizing dense optical flow estimation or complex attention mechanisms to maximize information extraction from multiple frames and ensure correct alignment before reconstruction.

However, achieving this impressive visual quality imposes a heavy demand on system resources due to the reliance on explicit motion compensation and deep feature extraction. While manageable on high-end GPUs, these architectures require memory and processing power that far exceed the capabilities of typical edge devices. Consequently, when deployed on computationally constrained platforms, the execution overhead exceeds real-time limits, rendering them impractical for applications where efficient processing is just as critical as restoration quality.

To address these problems, we propose NanoVSR, a lightweight video super-resolution network explicitly optimized for edge platforms like the NVIDIA Jetson. In the broader context of on-device AI, where real-time processing and strict resource constraints are paramount~\cite{wang2025empowering}, our architecture draws inspiration from highly efficient, compute-bound architectures~\cite{sandler2018mobilenetv2, ma2018shufflenetv2practicalguidelines, howard2019searchingmobilenetv3, tan2019efficientnetrethinkingmodelscaling}. We introduce a purely convolutional, bidirectional recurrent framework built upon structural reparameterization. This allows NanoVSR to train as a multi-branch network and mathematically collapse into an optimized stream of standard convolutions during inference. Furthermore, we entirely discard computationally expensive optical flow and traditional feature concatenation, opting instead for a direct additive propagation scheme. To compensate for the reduced architectural capacity and the lack of explicit alignment modules, we employ a progressive two-stage training method. The network is initially pre-trained on short video sequences to establish spatial feature extraction, and subsequently fine-tuned on extended sequences. This forces the bidirectional recurrent structure to implicitly learn residual motion and long-term dependencies, providing a solution that tightly balances real-time edge performance with the demand for high-quality video enhancement.

Our main contributions are as follows:
\begin{itemize}
    \item We propose NanoVSR, a highly efficient, purely convolutional VSR architecture explicitly designed for edge devices, eliminating the reliance on memory-intensive explicit alignment and custom CUDA operations.
    \item We introduce a direct additive propagation scheme for bidirectional recurrent VSR that avoids channel concatenation, reducing memory bandwidth and enabling the full network - including temporal propagation - to collapse into a plain convolution stream via structural reparameterization.
    \item We conduct extensive experiments across standard benchmarks (REDS4, Vid4, Vimeo90K-T) and provide hardware-specific profiling on two NVIDIA Jetson Orin NX configurations (8GB/15W and 16GB/25W), demonstrating that NanoVSR delivers a highly competitive balance of restoration quality and inference throughput compared to existing baselines.
\end{itemize}

\section{Related Work}
\label{sec:related}
In this section, we review the literature relevant to our proposed NanoVSR architecture. We first discuss the datasets commonly used for training and evaluating video super-resolution models, highlighting the shift towards more complex motion profiles. Subsequently, we provide an overview of existing VSR methodologies, ranging from traditional sliding-window networks to modern, computationally intensive transformers. Finally, we situate our work within the context of efficient network design, specifically focusing on structural reparameterization and progressive training strategies.

\subsection{Video Super-Resolution Datasets}
The evaluation of Video Super-Resolution (VSR) algorithms heavily depends on datasets that accurately reflect real-world degradation and complex temporal dynamics. Early works primarily relied on the Vid4~\cite{liu2014bayesian} and SPMCS~\cite{tao2017detail} datasets. However, these benchmarks lack the diversity and motion complexity required to test the limits of modern deep learning architectures. 

To bridge this gap, larger and more challenging datasets have become the standard for both training and robust evaluation. Vimeo-90K~\cite{xue2019video} provides over 90,000 high-resolution video clips with diverse scenes, serving as a primary training ground for most VSR networks. Concurrently, the REDS dataset~\cite{nah2019ntire}, introduced during the NTIRE 2019 Challenge, offers complex sequences degraded by heavy motion blur and large displacements. In this paper, we establish a rigorous evaluation protocol by benchmarking our NanoVSR across three widely adopted test sets: the classic Vid4 for historical comparisons, the Vimeo-90K test set (often denoted as Vimeo-90K-T) for diverse motion profiles, and the challenging REDS4 split (comprising clips 000, 011, 015, and 020) for highly dynamic, complex scenes. This comprehensive testing ensures a fair, transparent, and realistic assessment of the critical trade-off between restoration quality and edge-device inference time.

\subsection{Existing VSR Methods}
VSR architectures have transitioned from traditional sliding-window CNNs to recurrent frameworks and, more recently, heavy attention-based transformers.

\textbf{Sliding-Window and Alignment-Based CNNs.} 
Early VSR models processed a local temporal window to reconstruct a central frame. Methods like VESPCN~\cite{caballero2017real} and TOFlow~\cite{xue2019video} relied on explicit optical flow. To mitigate flow estimation errors, TDAN~\cite{tian2020tdan} and EDVR~\cite{wang2019edvr} introduced Deformable Convolutions (DCNs)~\cite{dai2017deformableconvolutionalnetworks, zhu2019deformableconvnetsv2deformable}. However, computing dense optical flow or DCN offsets for every frame imposes a severe computational bottleneck, rendering these methods unsuitable for real-time edge deployment.

\textbf{Recurrent Architectures.} 
Recurrent frameworks propagate latent features across the video sequence. Early models like FRVSR~\cite{sajjadi2018frame} and RLSP~\cite{fuoli2019efficient} struggled with long-term dependencies. This was largely overcome by BasicVSR~\cite{chan2021basicvsr} and its variants~\cite{chan2022basicvsr++, chan2021basicvsr}, which use bidirectional propagation and flow-based alignment. While more efficient, their reliance on sequential flow estimation (\eg, SPyNet~\cite{ranjan2017optical}) leads to suboptimal inference speeds on constrained hardware.

\textbf{Transformer-Based VSR.} 
Vision transformers currently dominate VSR leaderboards. VRT~\cite{liang2024vrt} extracts features using mutual attention, while RVRT~\cite{liang2022recurrent} employs guided self-attention. While RVRT sets the state-of-the-art (SOTA) in restoration quality, its sheer parameter count and quadratic attention complexity make it a computationally prohibitive model strictly confined to high-end GPUs.

\textbf{Efficient VSR and Structural Reparameterization.} 
Developing fast architectures for VSR remains underexplored due to the computational burden of temporal alignment. Existing attempts to streamline VSR by scaling down baselines (\eg, EDVR~\cite{wang2019edvr}), employing lighter attention (\eg, RVRT~\cite{liang2022recurrent}), or using recurrent designs (\eg, RSDN~\cite{isobe2020video}) still struggle to meet strict real-time constraints on embedded hardware. To address this, we draw inspiration from structural reparameterization~\cite{ding2021repvgg}, decoupling a rich multi-branch training topology from a minimalist single-branch inference architecture.

\section{Methodology}
\label{sec:methodology}

In this section, we detail the structural design and optimization strategy of NanoVSR, our lightweight video super-resolution network. We first introduce the overall bidirectional recurrent architecture, designed to maximize inference throughput by entirely bypassing explicit motion compensation and custom CUDA operations (\cref{subsec:architecture}). Next, we delve into the core of our efficiency paradigm: the structural reparameterization technique~\cite{ding2021repvgg} used to collapse a multi-branch training topology into a minimalist, single-branch inference model (\cref{subsec:reparam}). Finally, we describe the two-stage training protocol that enables our network to implicitly learn robust spatio-temporal alignments and achieve competitive restoration quality (\cref{subsec:training}).

\subsection{Architecture Overview}
\label{subsec:architecture}
Unlike recent heavyweight Video Super-Resolution (VSR) models that rely on explicit optical flow~\cite{sajjadi2018frame, xue2019video, wang2019edvr, chan2021basicvsr} or computationally expensive transformer-based attention mechanisms~\cite{liang2022recurrent, liang2024vrt, shi2022rethinking, cao2023video, dosovitskiy2021imageworth16x16words}, NanoVSR is designed as a purely convolutional, bidirectional recurrent network constructed exclusively from standard tensor operations. By explicitly avoiding custom CUDA modules like deformable convolutions and memory-intensive warping functions, NanoVSR ensures native ONNX~\cite{bai2017onnx} compatibility. This maximizes data throughput on edge devices by fully utilizing deep fusion optimizations provided by hardware accelerators (\eg, TensorRT~\cite{nvidia2017tensorrt}).

\begin{figure}[t]
  \centering
  \includegraphics[width=0.95\textwidth]{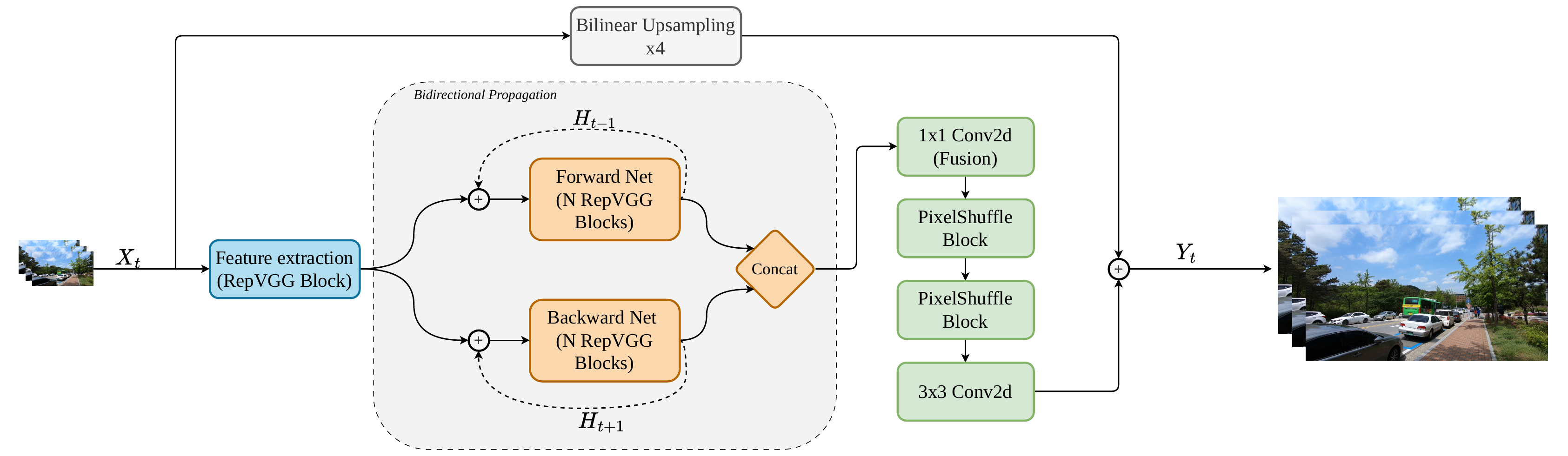}
  \caption{Detailed schematic of the NanoVSR architecture. LR input frames are embedded into a latent space and passed through a bidirectional recurrent structure built with reparameterizable blocks (see \cref{fig:reparam_scheme}). Crucially, to minimize memory bandwidth, temporal propagation relies on element-wise addition instead of standard feature concatenation. The refined features are bottlenecked and upsampled, with a global bilinear residual added to reconstruct the final HR sequence.}
  \label{fig:nanovsr_arch}
\end{figure}

The overall pipeline of NanoVSR is illustrated in \cref{fig:nanovsr_arch}. Given a sequence of low-resolution (LR) input frames $x \in \mathbb{R}^{T \times 3 \times H \times W}$, where $T$ denotes the sequence length, the network processes the sequence through three main stages: shallow feature extraction, bidirectional temporal propagation, and high-resolution (HR) reconstruction.

\textbf{Shallow Feature Extraction.}
Initially, the temporal dimension is folded into the batch dimension to process all frames independently and efficiently. A single feature extraction block maps the RGB input into a higher-dimensional latent space:
\begin{equation}
    f_i = \mathcal{H}_{ext}(x_i), \quad \forall i \in \{1, \dots, T\},
\end{equation}
where $f_i \in \mathbb{R}^{C \times H \times W}$ represents the extracted features for the $i$-th frame, $C$ is the number of feature channels, and $\mathcal{H}_{ext}$ denotes our multi-branch extraction block (which will be detailed in \cref{subsec:reparam}). After this operation, the features are reshaped back to restore the temporal dimension.

\textbf{Bidirectional Temporal Propagation.}
To aggregate temporal context without explicit motion compensation, we employ a bidirectional recurrent structure. In standard recurrent architectures~\cite{sajjadi2018frame, fuoli2019efficient, isobe2020video, chan2021basicvsr}, hidden states are typically concatenated with input features. However, concatenation doubles the channel dimension, leading to a substantial increase in computational cost and heavily burdening memory bandwidth during subsequent convolutions. To mitigate this, NanoVSR employs a direct element-wise additive propagation scheme. The forward and backward hidden states, denoted as $h^{\rightarrow}_i$ and $h^{\leftarrow}_i$ respectively, are updated sequentially as follows:
\begin{align}
    h^{\rightarrow}_i &= \mathcal{H}_{forward}\left(f_i + h^{\rightarrow}_{i-1}\right), \\
    h^{\leftarrow}_i &= \mathcal{H}_{backward}\left(f_i + h^{\leftarrow}_{i+1}\right),
\end{align}
where both $\mathcal{H}_{forward}$ and $\mathcal{H}_{backward}$ consist of a sequence of structural blocks configured with the LeakyReLU activation function. This additive formulation forces the network to implicitly learn residual motion representations, significantly reducing the memory footprint during inference while maintaining robust temporal coherence. The initial hidden states $h^{\rightarrow}_0$ and $h^{\leftarrow}_{T+1}$ are initialized as zero tensors.

\textbf{Fusion and Reconstruction.}
For each time step $i$, the forward and backward hidden states are channel-wise concatenated and fused via a $1 \times 1$ convolution to bottleneck the channel dimension back to $C$:
\begin{equation}
    f^{fused}_i = \mathcal{H}_{fusion}\left( [h^{\rightarrow}_i, h^{\leftarrow}_i] \right),
\end{equation}
where $[\cdot, \cdot]$ denotes the channel-wise concatenation operator. The fused features are subsequently upsampled through a cascaded sub-pixel convolution module (PixelShuffle)~\cite{shi2016real} consisting of two progressive $2\times$ spatial expansion stages. These stages are separated by PReLU~\cite{he2015delving} activations, before a final $3 \times 3$ convolution projects the refined representations back into the 3-channel RGB domain.

To ease the learning process, preserve absolute spatial scale, and allow the core network to focus strictly on high-frequency detail synthesis, we employ a global residual connection. The final HR output $\hat{y}_i$ is generated by adding the network's processed residual output to the upsampled LR input. To ensure maximum hardware efficiency and minimize latency, we substitute traditional bicubic upsampling with highly optimized bilinear interpolation for this base projection:
\begin{equation}
    \hat{y}_i = \mathcal{H}_{up}(f^{fused}_i) + \text{Bilinear}_{\uparrow 4}(x_i).
\end{equation}

\subsection{Structural Reparameterization for Inference}
\label{subsec:reparam}

The fundamental bottleneck for neural network deployment on edge devices is not merely the parameter count, but the memory access cost and network fragmentation, which severely limit hardware utilization~\cite{ma2018shufflenetv2practicalguidelines}. While multi-branch architectures provide strong representational capacity and stable gradient flow during training~\cite{he2016deepresiduallearningimage, szegedy2015goingdeeperconvolutions}, they fragment memory access and introduce severe CUDA kernel launch overheads during inference. To resolve this disparity, NanoVSR is constructed utilizing structural reparameterization, allowing us to completely decouple the training topology from the deployment architecture.

During the training phase, every fundamental building block operates as a multi-branch topology. The input features are processed via three parallel pathways: a $3 \times 3$ convolution, a $1 \times 1$ convolution, and an identity mapping, each immediately followed by a Batch Normalization (BN) layer~\cite{ioffe2015batchnormalizationacceleratingdeep}.

Upon completing the training phase, the model is prepared for hardware deployment. Because convolution and batch normalization act as linear and affine transformations during inference, we apply a mathematically equivalent transformation to collapse the entire block. Following the standard reparameterization formulation, the parallel branches and their respective absorbed BN statistics are zero-padded, spatially aligned, and linearly fused into a single set of weights and biases.

\begin{figure}[tb]
  \centering
  \includegraphics[width=0.95\textwidth]{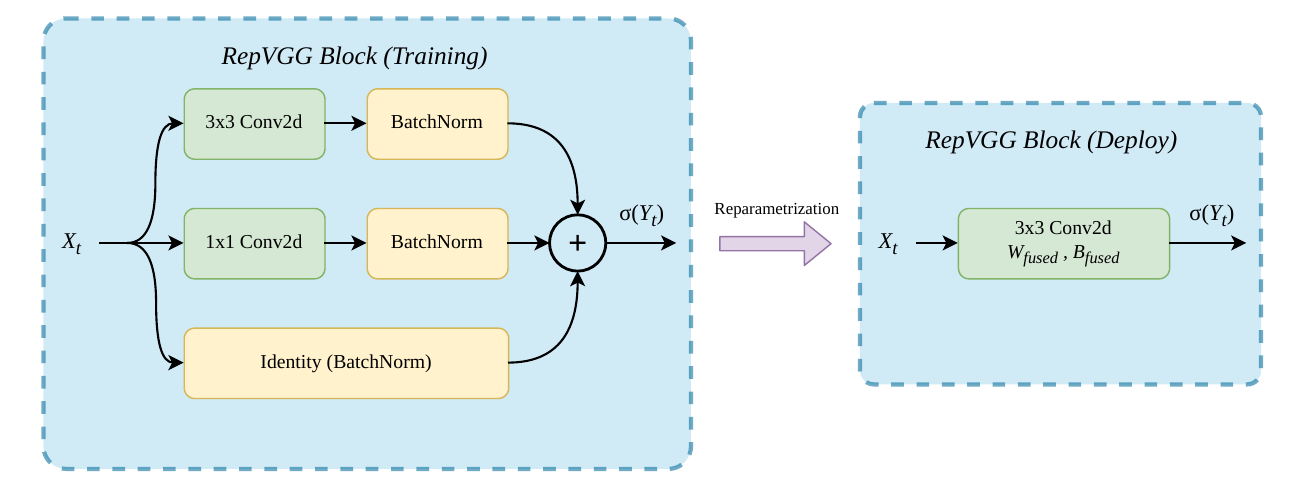} 
  \caption{Diagram of the structural reparameterization process~\cite{ding2021repvgg}. During training, the block utilizes a multi-branch topology ($3 \times 3$, $1 \times 1$, and identity pathways) to ensure robust feature extraction. For deployment, Batch Normalization statistics are first absorbed into their respective kernels. Finally, the parallel branches are mathematically fused into a single, dense $3 \times 3$ convolution, eliminating memory fragmentation and significantly accelerating inference.}
  \label{fig:reparam_scheme}
\end{figure}

Consequently, the complex multi-branch topology completely vanishes during edge deployment, collapsing into a single, highly optimized $3 \times 3$ convolution, as visually depicted in \cref{fig:reparam_scheme}. The inference forward pass is executed as a continuous stream of plain convolutions, completely eliminating memory fragmentation and making it perfectly suited for deep fusion optimizations provided by accelerators like TensorRT~\cite{nvidia2017tensorrt}.

\subsection{Two-Stage Training Strategy}
\label{subsec:training}

Training an implicit-alignment VSR network from scratch on long temporal sequences is notoriously unstable. To prevent gradient instability and ensure convergence, we adopt a two-phase progressive curriculum strategy~\cite{wang2022surveycurriculumlearning}.

\textbf{Phase 1: Short-Sequence Pre-training.}
For the initial 50,000 iterations, the network is trained on 7-frame sequences from the Vimeo-90K dataset. This phase focuses on establishing robust spatial feature extraction and short-term temporal fusion. The limited motion in these clips prevents early gradient instability, providing a solid initialization for the structural blocks.

\textbf{Phase 2: Long-Sequence Fine-tuning.}
Subsequently, we seamlessly switch the training corpus to the REDS dataset and expand the temporal window to 30 continuous frames for the remaining 100,000 iterations. We employ a sliding window approach to sample these sequences. This phase is crucial for developing the network's long-term recurrence capabilities, allowing the bidirectional hidden states to propagate information across complex motion trajectories.

\textbf{Optimization Details.}
The network is trained for 150,000 iterations using a global batch size of 12 and spatial patches of size $256 \times 256$. We optimize the model end-to-end using the Charbonnier penalty~\cite{charbonnier1994two} ($\epsilon = 1 \times 10^{-6}$) and the Adam optimizer~\cite{kingma2015adam} ($\beta_1=0.9, \beta_2=0.99$). A single Cosine Annealing scheduler~\cite{loshchilov2017sgdr} governs the learning rate across both phases, decaying from $3 \times 10^{-4}$ to $1 \times 10^{-7}$. To maintain high throughput and memory efficiency, we utilize Automatic Mixed Precision (AMP)~\cite{micikevicius2018mixed} with the BFloat16 format and gradient clipping with a maximum $L_2$ norm of 0.5 throughout the training process. Data augmentation includes geometric transformations (flips, rotations, temporal reversal) and the CutBlur strategy~\cite{yoo2020rethinking}.

\section{Experiments}
\label{sec:experiments}

This section evaluates NanoVSR across several standard benchmarks. Our analysis focuses on two primary dimensions: image reconstruction fidelity, quantified by PSNR and SSIM metrics, and hardware-specific efficiency on edge devices.

\subsection{Experimental Setup}
\label{subsec:setup}

The models are trained according to the progressive curriculum strategy detailed in \cref{subsec:training} using the Vimeo-90K and REDS datasets. To ensure a rigorous and unbiased evaluation, the four sequences (000, 011, 015, 020) that comprise the REDS4 benchmark were strictly excluded from the training set. Performance is evaluated on three standard benchmarks: REDS4, Vid4, and Vimeo90K-T.

In accordance with established video super-resolution literature, for the REDS4 dataset, we report Peak Signal-to-Noise Ratio (PSNR) and Structural Similarity Index (SSIM)~\cite{wang2004image} metrics calculated in the RGB color space. For the Vid4 and Vimeo90K-T benchmarks, metrics are computed on the luminance channel (Y-channel) after converting the network's RGB output. All models are evaluated at a spatial upsampling factor of $4\times$.

Quantitative evaluation and inference benchmarks are conducted on a single NVIDIA H100 GPU. To ensure a fair and direct comparison, the inference latency for all baseline methods was re-evaluated on our hardware using FP32 precision. For the edge deployment analysis, we utilize two configurations of the NVIDIA Jetson Orin NX: an 8GB variant restricted to a 15W TDP and a 16GB variant operating at a 25W TDP. All models are exported via ONNX and compiled into optimized execution engines using NVIDIA TensorRT 10.3.0.30~\cite{nvidia2017tensorrt}. To maximize performance on these edge platforms, we employ FP16 precision and a chunk-based execution pattern with a temporal window of 15 frames ($T=15$), ensuring efficient bidirectional context aggregation within the devices memory and power constraints.

\subsection{Quantitative Results}
\label{subsec:quantitative}

\begin{table*}[tb]
\caption{Quantitative comparison on Vid4, REDS4, and Vimeo90K-T. Runtime is measured in ms/frame. Metrics are reported as PSNR (dB) / SSIM. For baseline methods, results on the REDS4~\cite{nah2019ntire} benchmark reflect models trained on the REDS dataset, whereas results on Vid4~\cite{liu2014bayesian} and Vimeo90K-T~\cite{xue2019video} correspond to models trained on Vimeo-90K. These separate evaluations have been merged into this single table.}
\label{tab:quantitative}
\centering
\resizebox{\textwidth}{!}{
\setlength{\tabcolsep}{7.5pt}
\begin{tabular}{@{}l c c c c c@{}}\toprule
\textbf{Model} & \textbf{Params} & \textbf{Runtime} & \textbf{Vid4~\cite{liu2014bayesian}} & \textbf{REDS4~\cite{nah2019ntire}} & \textbf{Vimeo90K-T~\cite{xue2019video}} \\
 & & \textbf{(ms)} & \scriptsize (Y-channel) & \scriptsize (RGB) & \scriptsize (Y-channel) \\
\midrule
NanoVSR-226k & 226k & 1.910 & 25.26 / 0.7252 & 28.23 / 0.8057 & 34.31 / 0.9130 \\
NanoVSR-644k & 644k & 2.982 & 26.05 / 0.7761 & 28.64 / 0.8215 & 35.00 / 0.9226 \\
NanoVSR-1.7M & 1.7M & 4.268 & 26.44 / 0.7964 & 29.15 / 0.8364 & 35.49 / 0.9294 \\
NanoVSR-5.4M & 5.4M & 8.547 & 26.76 / 0.8089 & 29.73 / 0.8526 & 35.85 / 0.9335 \\
\midrule
EDVR-M~\cite{wang2019edvr} & 3.3M & 27.343 & 27.10 / 0.8186 & 30.53 / 0.8699 & 37.09 / 0.9446 \\
BasicVSR~\cite{chan2021basicvsr} & 6.2M & 15.160 & 27.24 / 0.8251 & 31.42 / 0.8909 & 37.18 / 0.9450 \\
IconVSR~\cite{chan2021basicvsr} & 8.7M & 28.881 & 27.39 / 0.8279 & 31.67 / 0.8948 & 37.47 / 0.9476 \\
RVRT~\cite{liang2022recurrent} & 10.8M & 47.867 & 27.99 / 0.8462 & 32.75 / 0.9113 & 38.15 / 0.9527 \\
EDVR-L~\cite{wang2019edvr} & 20.6M & 73.994 & 27.35 / 0.8264 & 31.09 / 0.8800 & 37.61 / 0.9489 \\
\bottomrule
\end{tabular}
}
\end{table*}

We evaluate NanoVSR against several state-of-the-art VSR baselines on the Vid4, REDS4, and Vimeo90K-T benchmarks. Our analysis considers both restoration quality (PSNR, SSIM) and computational efficiency, including parameter count and runtime per frame for a $180 \times 320$ input. 

\begin{figure}[t]
    \centering
    \setlength{\fboxrule}{0.6pt}
    \setlength{\fboxsep}{-\fboxrule}

    \begin{minipage}[b]{0.23\textwidth}
        \begin{subfigure}[b]{1.00\textwidth}
            \centering
            \includegraphics[width=\textwidth]{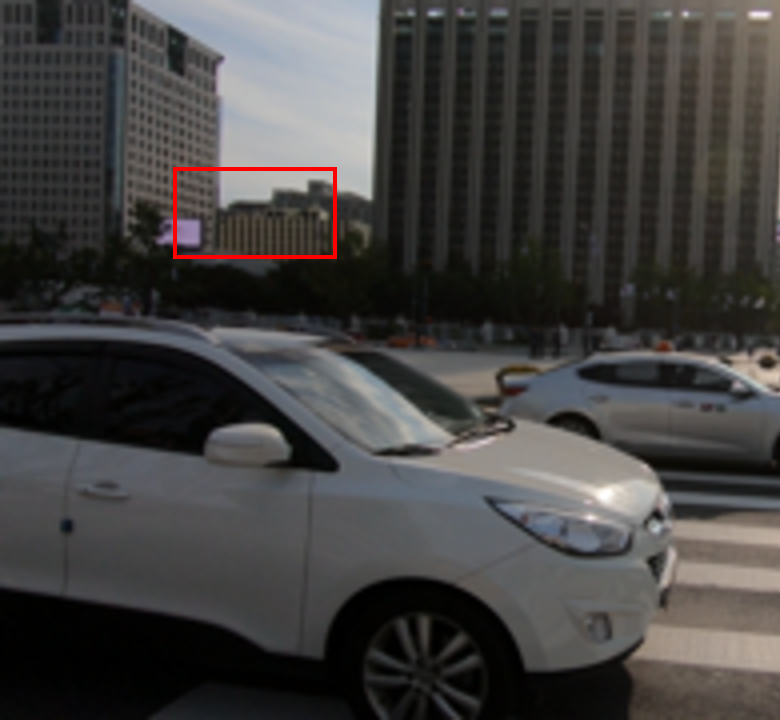}
            \caption*{Frame 055, Clip 015}
        \end{subfigure}
    \end{minipage}
    \hfill
    \begin{minipage}[b]{0.76\textwidth}
        \centering
        
        \begin{subfigure}[b]{0.24\textwidth}
            \centering
            \caption*{LR(x4)}
            \color{red}
            \fbox{\includegraphics[width=\textwidth]{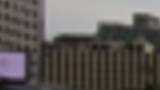}}
        \end{subfigure}
        \hfill
        \begin{subfigure}[b]{0.24\textwidth}
            \centering
            \caption*{BasicVSR~\cite{chan2021basicvsr}}
            \color{red}
            \fbox{\includegraphics[width=\textwidth]{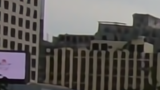}}
        \end{subfigure}
        \hfill
        \begin{subfigure}[b]{0.24\textwidth}
            \centering
            \caption*{EDVR-M~\cite{wang2019edvr}}
            \color{red}
            \fbox{\includegraphics[width=\textwidth]{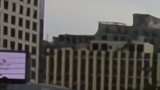}}
        \end{subfigure}
        \hfill
        \begin{subfigure}[b]{0.24\textwidth}
            \centering
            \caption*{RVRT~\cite{liang2022recurrent}}
            \color{red}
            \fbox{\includegraphics[width=\textwidth]{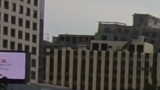}}
        \end{subfigure}

        \begin{subfigure}[b]{0.24\textwidth}
            \centering
            \color{red}
            \fbox{\includegraphics[width=\textwidth]{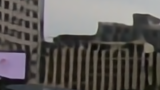}}
            \caption*{\textbf{NanoVSR-644k}}
        \end{subfigure}
        \hfill
        \begin{subfigure}[b]{0.24\textwidth}
            \centering
            \color{red}
            \fbox{\includegraphics[width=\textwidth]{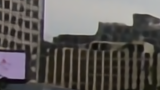}}
            \caption*{\textbf{NanoVSR-1.7M}}
        \end{subfigure}
        \hfill
        \begin{subfigure}[b]{0.24\textwidth}
            \centering
            \color{red}
            \fbox{\includegraphics[width=\textwidth]{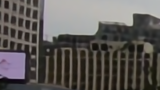}}
            \caption*{\textbf{NanoVSR-5.4M}}
        \end{subfigure}
        \hfill
        \begin{subfigure}[b]{0.24\textwidth}
            \centering
            \color{red}
            \fbox{\includegraphics[width=\textwidth]{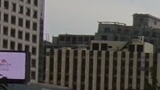}}
            \caption*{GT}
        \end{subfigure}
    \end{minipage}

    \caption{Qualitative comparison on REDS4~\cite{nah2019ntire} dataset. Left: full frame, Right: magnified patches.}
    \label{fig:qualitative_reds4}
\end{figure}

\begin{figure}[t]
    \centering
    \setlength{\fboxrule}{0.6pt}
    \setlength{\fboxsep}{-\fboxrule}

    \begin{minipage}[b]{0.23\textwidth}
        \begin{subfigure}[b]{1.00\textwidth}
            \centering
            \includegraphics[width=\textwidth]{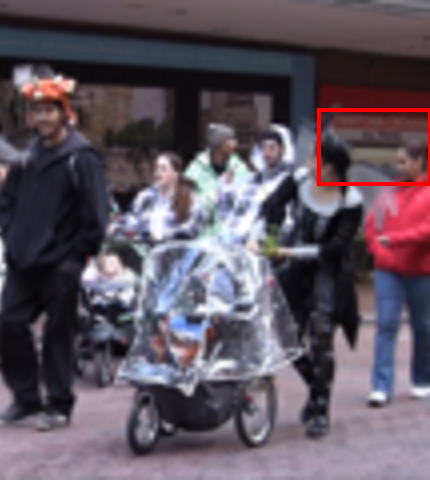}
            \caption*{Frame 038, Clip walk}
        \end{subfigure}
    \end{minipage}
    \hfill
    \begin{minipage}[b]{0.76\textwidth}
        \centering
        
        \begin{subfigure}[b]{0.24\textwidth}
            \centering
            \caption*{LR(x4)}
            \color{red}
            \fbox{\includegraphics[width=\textwidth]{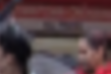}}
        \end{subfigure}
        \hfill
        \begin{subfigure}[b]{0.24\textwidth}
            \centering
            \caption*{BasicVSR~\cite{chan2021basicvsr}}
            \color{red}
            \fbox{\includegraphics[width=\textwidth]{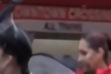}}
        \end{subfigure}
        \hfill
        \begin{subfigure}[b]{0.24\textwidth}
            \centering
            \caption*{EDVR-M~\cite{wang2019edvr}}
            \color{red}
            \fbox{\includegraphics[width=\textwidth]{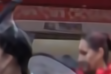}}
        \end{subfigure}
        \hfill
        \begin{subfigure}[b]{0.24\textwidth}
            \centering
            \caption*{RVRT~\cite{liang2022recurrent}}
            \color{red}
            \fbox{\includegraphics[width=\textwidth]{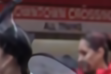}}
        \end{subfigure}

        \begin{subfigure}[b]{0.24\textwidth}
            \centering
            \color{red}
            \fbox{\includegraphics[width=\textwidth]{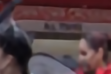}}
            \caption*{\textbf{NanoVSR-644k}}
        \end{subfigure}
        \hfill
        \begin{subfigure}[b]{0.24\textwidth}
            \centering
            \color{red}
            \fbox{\includegraphics[width=\textwidth]{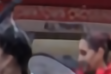}}
            \caption*{\textbf{NanoVSR-1.7M}}
        \end{subfigure}
        \hfill
        \begin{subfigure}[b]{0.24\textwidth}
            \centering
            \color{red}
            \fbox{\includegraphics[width=\textwidth]{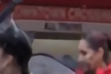}}
            \caption*{\textbf{NanoVSR-5.4M}}
        \end{subfigure}
        \hfill
        \begin{subfigure}[b]{0.24\textwidth}
            \centering
            \color{red}
            \fbox{\includegraphics[width=\textwidth]{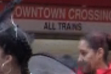}}
            \caption*{GT}
        \end{subfigure}
    \end{minipage}

    \caption{Qualitative comparison on Vid4~\cite{liu2014bayesian} dataset. Left: full frame, Right: magnified patches.}
    \label{fig:qualitative_vid4}
\end{figure}

As detailed in \cref{tab:quantitative}, NanoVSR exhibits a significant speed advantage over existing methods, delivering high performance while maintaining sharp reconstruction quality. While heavyweight transformer models like RVRT~\cite{liang2022recurrent} achieve the highest PSNR, they incur substantial inference delay ($\sim 48$ ms), making them unsuitable for real-time applications on edge devices. In contrast, our ultra-lightweight variant, NanoVSR-226k, operates with an inference time of just 1.91 ms, providing remarkable efficiency gains. Striking the optimal balance between speed and quality, our flagship NanoVSR-644k baseline achieves 28.64 dB on REDS4 requiring only 2.98 ms per frame, directly rivaling significantly heavier architectures. Furthermore, scaling up to the NanoVSR-1.7M variant yields a competitive 29.15 dB PSNR on the challenging REDS4 benchmark while still maintaining a low runtime of 4.268 ms per frame. This demonstrates that NanoVSR can deliver solid restoration quality at a fraction of the computational cost of traditional architectures. Qualitative results for the REDS4 dataset are illustrated in \cref{fig:qualitative_reds4}, demonstrating that our models effectively recover sharp details comparable to much larger architectures. Visual comparisons on the Vid4 benchmark, specifically for text-heavy scenes (see \cref{fig:qualitative_vid4}), reveal that NanoVSR performs on par with significantly heavier models. In this context, both lightweight and state-of-the-art models exhibit only average restoration quality, highlighting the persistent challenge of alpha-numeric reconstruction in motion-heavy sequences.

\subsection{Efficiency Analysis}

To demonstrate the real-world applicability and scalability of our method on embedded hardware, we conduct an extensive efficiency analysis across two configurations of the NVIDIA Jetson Orin NX: an 8GB variant restricted to a 15W TDP and a 16GB variant operating at a 25W TDP. Our evaluation focuses on inference latency and throughput (FPS) for processing a temporal window of $T=15$ frames at two different input resolutions: $180 \times 320$ and $270 \times 480$. All models are compiled and optimized using the TensorRT 10.3 execution engine. Since both devices were operated at their maximum performance profiles and memory consumption remained consistently within hardware limits due to static engine optimization, we prioritize throughput as the primary metric for edge-device viability.

The results, summarized in \cref{tab:efficiency_comparison}, clearly validate our hardware-aware design principles. Heavyweight architectures, represented here by BasicVSR~\cite{chan2021basicvsr}, struggle to maintain real-time performance on edge devices, achieving not even 4 FPS for a $270 \times 480$ input sequence. Furthermore, they demand significantly more resources, which poses a severe bottleneck on embedded platforms.

In contrast, NanoVSR transforms the VSR task into a highly optimized process. Our ultra-lightweight variant, NanoVSR-226k, meets stringent streaming constraints, delivering almost 44 FPS on $180 \times 320$ inputs and nearly 20 FPS on $270 \times 480$ inputs on Jetson Orin NX (16GB). Delivering an exceptional efficiency-to-quality ratio, our flagship baseline, NanoVSR-644k, achieves a remarkable $\sim$27 FPS on $180 \times 320$ inputs. Even our scaled NanoVSR-1.7M model maintains competitive edge performance, demonstrating that our architecture scales elegantly without exceeding the strict hardware limits inherent to edge computing.

\begin{table}[t]
\caption{Efficiency scalability across NVIDIA Jetson Orin NX configurations. We compare average inference latency per frame (ms) and throughput (FPS) between the 8GB variant (15W TDP) and the 16GB variant (25W TDP). All models are compiled via TensorRT with a temporal window of $T=15$.}
\label{tab:efficiency_comparison}
\centering
\resizebox{\textwidth}{!}{
\setlength{\tabcolsep}{8pt}
\begin{tabular}{@{}l c cc cc@{}}
\toprule
 & & \multicolumn{2}{c}{\textbf{Orin NX (8GB)}} & \multicolumn{2}{c}{\textbf{Orin NX (16GB)}} \\
\cmidrule(lr){3-4} \cmidrule(lr){5-6}
\textbf{Model} & \textbf{Resolution} & \textbf{Runtime} & \textbf{FPS} & \textbf{Runtime} & \textbf{FPS} \\
 & & \textbf{(ms)} & & \textbf{(ms)} & \\
\midrule
NanoVSR-226k & $180 \times 320$ & 42.46 & 23.55 & 22.80 & 43.86 \\
             & $270 \times 480$ & 95.15 & 10.51 & 51.14 & 19.55 \\
\midrule
NanoVSR-644k & $180 \times 320$ & 62.04 & 16.12 & 36.77 & 27.20 \\
             & $270 \times 480$ & 139.03 & 7.19 & 77.99 & 12.82 \\
\midrule
NanoVSR-1.7M & $180 \times 320$ & 86.46 & 11.57 & 51.06 & 19.58 \\
             & $270 \times 480$ & 196.40 & 5.09 & 112.77 & 8.87 \\
\midrule
NanoVSR-5.4M & $180 \times 320$ & 190.71 & 5.24 & 115.50 & 8.66 \\
             & $270 \times 480$ & 427.95 & 2.34 & 263.82 & 3.79 \\
\midrule
BasicVSR~\cite{chan2021basicvsr} & $180 \times 320$ & 229.28 & 4.36 & 124.37 & 8.04 \\
                                 & $270 \times 480$ & 528.38 & 1.89 & 289.13 & 3.46 \\
\bottomrule
\end{tabular}
}
\end{table}

\subsection{Model Scaling Analysis}
\label{subsec:model_scaling}

\begin{figure}[t]
  \centering
  \includegraphics[width=0.95\textwidth]{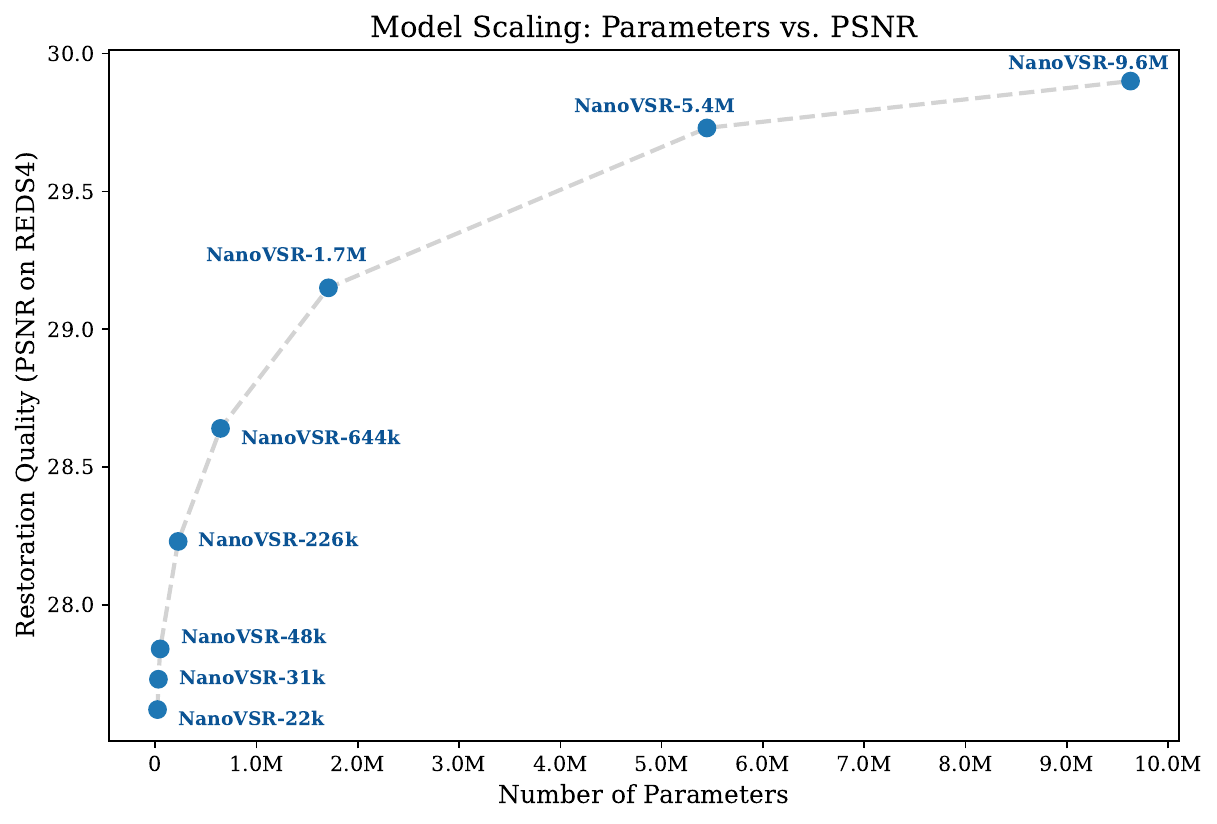}
  \caption{Model scaling analysis on the REDS4~\cite{nah2019ntire} dataset. The plot illustrates the trade-off between the number of parameters and restoration quality (PSNR). While scaling up the architecture consistently improves performance, the curve demonstrates diminishing returns beyond 5.4M parameters.}
  \label{fig:model_scaling}
\end{figure}

To evaluate the scalability of the proposed architecture, we investigate the relationship between the number of parameters and the restoration quality on the challenging REDS4 dataset. As illustrated in \cref{fig:model_scaling}, we scale the network capacity from an ultra-lightweight 22k parameters up to 9.6M parameters by progressively adjusting the number of temporal propagation blocks and internal feature channels.

\textbf{Initial Capacity Gains.}
The performance trajectory exhibits steep initial gains in the low-parameter regime. Scaling the network from the NanoVSR-22k variant to the NanoVSR-644k baseline yields a substantial improvement of 1.02 dB (from 27.62 dB to 28.64 dB). As the architectural capacity increases beyond the 1M parameter mark, the network continues to improve visual fidelity steadily, reaching 29.73 dB for the 5.4M variant.

\textbf{Performance Saturation.}
However, scaling the architecture from the Nano\-VSR-5.4M to the NanoVSR-9.6M variant yields diminishing returns. Despite a near two-fold increase in parameter count, it provides only a marginal PSNR improvement of 0.17 dB, effectively plateauing at 29.90 dB. This saturation point indicates that while the NanoVSR topology can easily scale to achieve higher-quality reconstructions, our baseline (NanoVSR-644k) strikes the most practical balance between computational efficiency and restoration quality, making it the definitive choice for practical edge deployment.

\subsection{Ablation Studies}
\label{subsec:ablation}

To explore the effectiveness of different architectural design choices, we conduct ablation studies on the REDS4, Vid4, and Vimeo90K-T datasets. Due to the substantial computational cost of video training, all ablations were conducted using our highly efficient NanoVSR-226k variant. We empirically observed that these design insights and relative performance improvements transfer consistently to our larger flagship model, NanoVSR-644k.

\begin{table}[tb]
\caption{Comprehensive ablation study on the NanoVSR architecture. We evaluate the impact of structural reparameterization (-NOFUSE, -SINGLE), curriculum pre-training (-NOPRET), and explicit motion alignment (-SPYNET). Runtime is measured in milliseconds per frame for a $180 \times 320$ input on an NVIDIA H100 GPU. Restoration quality is reported as PSNR (dB) / SSIM.}
  \label{tab:ablation}
    \centering
    \resizebox{\textwidth}{!}{
    \setlength{\tabcolsep}{7.5pt}
    \begin{tabular}{@{}l c c c c c@{}}
    \toprule
    \textbf{Model} & \textbf{Params} & \textbf{Runtime} & \textbf{Vid4~\cite{liu2014bayesian}} & \textbf{REDS4~\cite{nah2019ntire}} & \textbf{Vimeo90K-T~\cite{xue2019video}} \\
     & & \textbf{(ms)} & \scriptsize (Y-channel) & \scriptsize (RGB) & \scriptsize (Y-channel) \\
    \midrule
    NanoVSR-226k (Reference) & 226k & 1.910 & 25.26 / 0.7252 & 28.23 / 0.8057 & 34.31 / 0.9130 \\
    NanoVSR-226k-NOFUSE & 245k & 3.471 & 25.26 / 0.7252 & 28.23 / 0.8057 & 34.31 / 0.9130 \\
    NanoVSR-226k-SINGLE & 226k & 1.885 & 25.25 / 0.7296 & 28.13 / 0.8027 & 34.25 / 0.9134 \\
    NanoVSR-226k-NOPRET & 226k & 1.887 & 25.17 / 0.7271 & 28.23 / 0.8054 & 34.21 / 0.9130 \\
    NanoVSR-226k-SPYNET & 1.7M & 3.953 & 25.95 / 0.7701 & 29.44 / 0.8460 & 34.85 / 0.9220 \\
    \bottomrule
  \end{tabular}
  }
\end{table}

\textbf{The impact of structural reparameterization.} In NanoVSR, we utilize structural reparameterization to decouple the training topology from the inference architecture. As shown in \cref{tab:ablation}, the baseline NanoVSR-226k and its non-fused counterpart, NanoVSR-226k-NOFUSE, achieve mathematically identical restoration quality. However, executing the parameter fusion reduces the total parameter count from 245k to 226k and nearly halves the inference time (from 3.471 ms to 1.910 ms). Conversely, NanoVSR-226k-SINGLE, which is trained from scratch as a strict single-branch architecture, suffers a noticeable degradation in overall reconstruction fidelity, dropping by 0.1 dB PSNR on REDS4. While the lower-capacity SINGLE variant yields a marginally higher SSIM on Vid4, this is a known mathematical consequence of its constrained capacity. As reflected by the PSNR drop, the restricted single-branch topology biases the network toward lower-frequency predictions. While this smoothing effect can inadvertently inflate SSIM scores on specific datasets, the model lacks the representational power to reconstruct the fine, high-frequency structural details captured by our multi-branch reference architecture. This confirms that the rich multi-branch training phase is essential for learning robust representations.

\textbf{The impact of curriculum pre-training.} Our two-stage progressive training scheme is evaluated by comparing the Nano\-VSR-226k model against Nano\-VSR-226k-NOPRET, which is trained exclusively on REDS, omitting the Vimeo-90K phase. While both models reach an identical PSNR on REDS4, the pre-trained variant naturally performs better on Vimeo90K-T due to prior exposure. For zero-shot generalization on the external Vid4 dataset, skipping this initial phase leads to a PSNR drop from 25.26 dB to 25.17 dB. Although NanoVSR-226k-NOPRET yields a marginally higher SSIM here (0.7271 vs. 0.7252), the overall metrics demonstrate that initial exposure on diverse sequences improves overall generalization and restoration quality.

\textbf{The impact of explicit alignment.} We analyze the architectural trade-off of omitting explicit motion compensation by introducing NanoVSR-226k-SPYNET. This variant shares the identical core architecture and is trained under the same progressive curriculum, with the sole addition of an integrated SPyNet~\cite{ranjan2017optical} module. As indicated in \cref{tab:ablation}, explicit alignment significantly boosts the restoration fidelity, yielding a substantial gain of 1.21 dB over Nano\-VSR-226k on REDS4. However, this performance surge incurs severe computational penalties. The total parameter count inflates to over 1.7M (of which approximately 1.5M belong strictly to the SPyNet estimator), and the inference time more than doubles to 3.953 ms per frame. Given the strict power and inference speed constraints inherent to edge devices, the NanoVSR architecture intentionally foregoes explicit optical flow.

\section{Limitations and Societal Impact}
\label{sec:limitations_impact}

While NanoVSR achieves a highly competitive efficiency-quality balance, it inherits limitations inherent to implicit-alignment architectures. The absence of explicit optical flow means the network may occasionally struggle to reconstruct high-frequency textures in the presence of extreme, non-rigid local motions. Furthermore, because bidirectional propagation requires a temporal look-ahead window ($T=15$), the model introduces a slight buffering delay, making it unsuitable for strict frame-by-frame real-time applications without degrading to a unidirectional mode.

From a societal perspective, our focus on extreme computational efficiency democratizes access to high-quality video processing and significantly mitigates the environmental footprint of deploying deep models at scale~\cite{schwartz2020greenai, strubell2019energypolicyconsiderationsdeep}. It can also extend the usable lifecycle of legacy low-resolution camera hardware, reducing electronic waste. However, enhancing low-resolution footage is inherently dual-use and could inadvertently facilitate non-consensual surveillance. It is crucial to emphasize that super-resolution algorithms probabilistically synthesize missing high-frequency details rather than recovering true signals; therefore, outputs must be treated as plausible enhancements rather than factual, identifiable evidence. We advocate for deploying VSR technologies within bounded, consent-driven domains such as media restoration, video conferencing, and broadcasting, while establishing robust ethical guidelines against unverified monitoring.

\section{Conclusion}
\label{sec:conclusion}

In this work, we presented NanoVSR, a highly efficient and fully convolutional architecture tailored for rapid video super-resolution on edge devices with limited resources. By moving away from costly optical flow estimation and heavy attention mechanisms, we demonstrated that robust spatial and temporal alignments can be effectively learned through a progressive curriculum training strategy. Furthermore, the integration of structural reparameterization allows our bidirectional recurrent network to collapse into a streamlined sequence of standard convolutions during inference. This design choice ensures seamless hardware compatibility with accelerators like TensorRT and drastically minimizes execution latency.

Evaluations on standard benchmarks, including REDS4, Vid4, and Vimeo90K-T, confirm that NanoVSR successfully redefines the trade-off between accuracy and efficiency for compact models. Our ultra-lightweight variant, NanoVSR-226k, achieves unprecedented inference speeds, delivering 43.8 FPS on an NVIDIA Jetson Orin NX (25W). Simultaneously, our flagship NanoVSR-644k baseline maintains a real-time throughput of 27.2 FPS while providing superior restoration quality. For high-performance scenarios, the scaled NanoVSR-1.7M variant reaches 29.15~dB on REDS4 with an H100 processing time of just 4.27 ms per frame. By significantly reducing parameter count without compromising visual fidelity, NanoVSR offers a highly practical and scalable solution for local edge applications. We hope our optimized approach encourages further research into sustainable and globally accessible video enhancement technologies.

\newpage

\section*{Acknowledgements}
The calculations were carried out using the computers of the Centre of Informatics Tricity Academic Supercomputer \& Network.

The research was supported in part by project ``Cloud Artificial Intelligence Service Engineering (CAISE) platform to create universal and smart services for various application areas'', No. KPOD.05.10-IW.10-0005/24, as part of the European IPCEI-CIS program, financed by NRRP (National Recovery and Resilience Plan) funds.

\bibliographystyle{splncs04}
\bibliography{main}
\end{document}